\documentclass[conference]{IEEEtran}
\pdfoutput=1
\usepackage[T1]{fontenc}
\IEEEoverridecommandlockouts
\usepackage{cite}
\usepackage{amsmath,amssymb,amsfonts}
\usepackage{algorithmic}
\usepackage{graphicx}
\usepackage{textcomp}
\usepackage{multirow}
\usepackage{array}
\usepackage{listings}
\usepackage{tablefootnote}
\usepackage{subcaption}
\usepackage{arydshln}
\usepackage{booktabs}
\usepackage{makecell}
\usepackage{hyperref}
\usepackage[normalem]{ulem}
\useunder{\uline}{\ul}{}
\usepackage[inline]{enumitem}
\usepackage{enumitem}
\usepackage{xcolor}
\def\BibTeX{{\rm B\kern-.05em{\sc i\kern-.025em b}\kern-.08em
    T\kern-.1667em\lower.7ex\hbox{E}\kern-.125emX}}
\begin{document}

\title{Teaching LLMs for Step-Level Automatic Math Correction via Reinforcement Learning}

\author {
    Junsong Li\textsuperscript{\rm 1},
    Jie Zhou\textsuperscript{\rm 1*}\thanks{*Corresponding author, jzhou@cs.ecnu.edu.cn.}, 
    Yutao Yang\textsuperscript{\rm 1}, 
    Bihao Zhan\textsuperscript{\rm 1}, 
    Qianjun Pan\textsuperscript{\rm 1}, 
    Yuyang Ding\textsuperscript{\rm 1}\\
    Qin Chen\textsuperscript{\rm 1},
    Jiang Bo\textsuperscript{\rm 2}, 
    Xin Lin\textsuperscript{\rm 2}, 
    Liang He\textsuperscript{\rm 1} \\
    \textit{\textsuperscript{\rm 1}School of Computer Science and Technology, East China Normal University, Shanghai, China}\\
    \textit{\textsuperscript{\rm 2}Lab of AI for Education, East China Normal University, Shanghai, China}\\
    {\small \texttt{junsong-li@stu.ecnu.edu.cn}, \texttt{jzhou@cs.ecnu.edu.cn}}
}


\maketitle

\begin{abstract}
Automatic math correction aims to check students' solutions to mathematical problems via artificial intelligence technologies. Most existing studies focus on judging the final answer at the problem level, while they ignore detailed feedback on each step in a math problem-solving process, which requires abilities of semantic understanding and reasoning.
In this paper, we propose a reinforcement learning (RL)-based method to boost large language model (LLM) for step-level automatic math correction, named \texttt{StepAMC}.
Particularly, we convert the step-level automatic math correction within the text classification task into an RL problem to enhance the reasoning capabilities of LLMs. 
Then, we design a space-constrained policy network to improve the stability of RL. 
Then, we introduce a fine-grained reward network to convert the binary human feedback into a continuous value. 
We conduct extensive experiments over two benchmark datasets and the results show that our model outperforms the eleven strong baselines.
\end{abstract}


\section{Introduction}
\label{sec:intro}
In traditional classroom settings, the process of diagnosing students' errors in exercises largely relies on teachers' manual effort, which can be time-consuming, especially when faced with large volumes of assignments.
Recently, artificial intelligence (AI) has been increasingly applied in education, bringing transformative potential to various aspects, such as intelligent tutoring~\cite{anderson1985intelligent,graesser2012intelligent}, personalized learning plans~\cite{basham2016operationalized,bulger2016personalized}, and automated grading systems~\cite{aldriye2019automated,messer2024automated}. 
Among these, providing timely and accurate feedback on students' exercises plays a critical role in enhancing their learning efficiency and correcting misconceptions. In this paper, we focus on automatic math correction systems, which aim to correct the answers and solutions of the students \cite{sukkarieh2003automarking}. 

\begin{figure}[t!]
    \centering
    \includegraphics[width=0.98\linewidth]{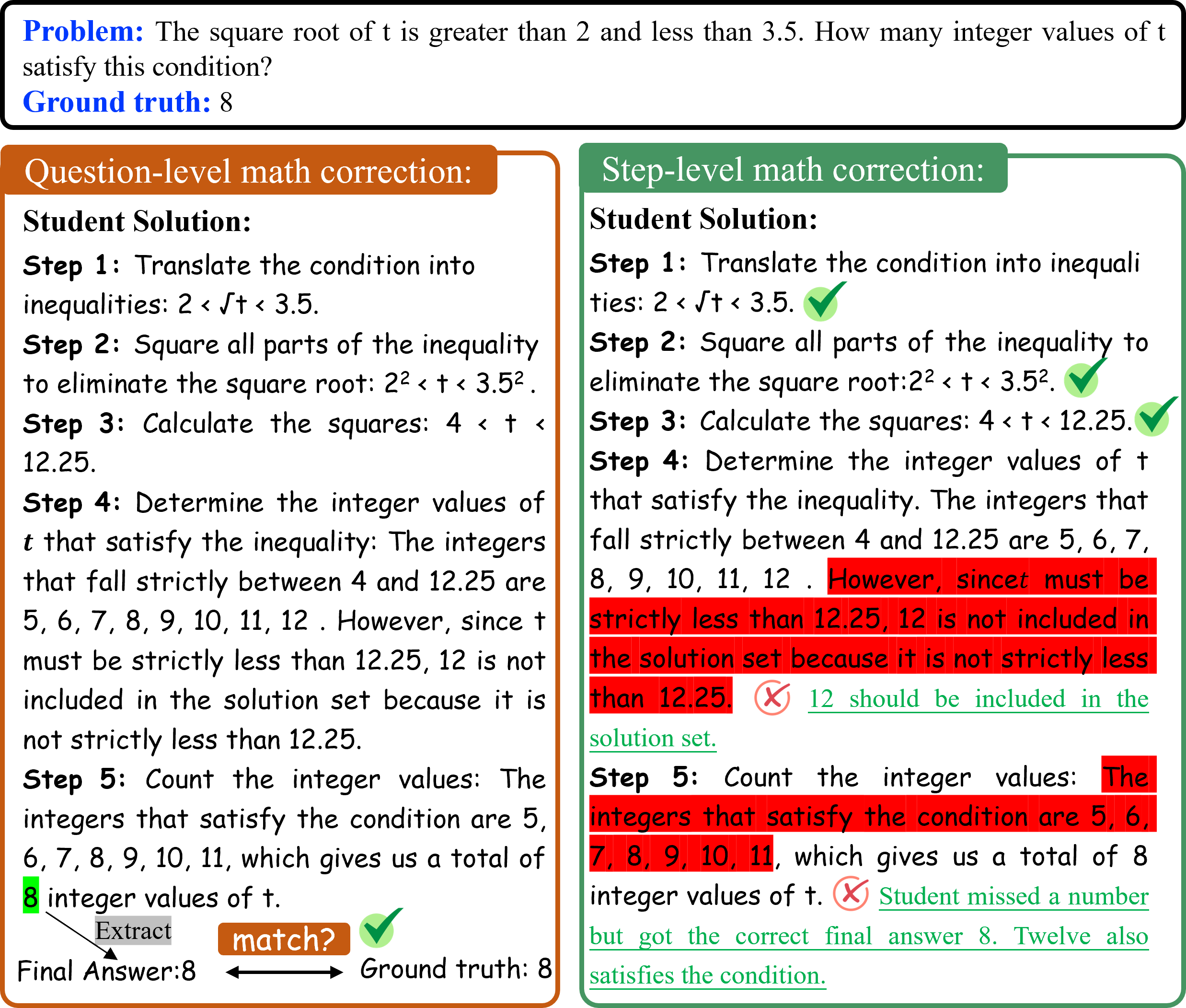}
    \caption{Examples of question- and step-level math correction.} 
    \label{fig:1} 
\vspace{-4mm}
\end{figure}

Prior work on automatic math correction has primarily focused on question-level tasks such as multiple-choice, fill-in-the-blank, and final answer verification \cite{suleman2008automatic,wang2018intelligent,li2019bags}. These methods extract the final answer from a learner’s solution and compare it to the ground truth, offering feedback based on correctness. While question-level correction is valuable for immediate feedback on final answers, it lacks the depth needed to evaluate the underlying reasoning that may lead to an incorrect answer. For example, the model extracts the answer ``8" from the solution and compares it with ground truth. However, the detailed steps in the solution may be wrong (Fig. \ref{fig:1}). We focus on step-level correction, which requires a more complex understanding of the problem-solving process, as it involves reasoning through the individual steps of a solution \cite{chaowicharat2023step}. This type of correction aims to not only assess the final answer but also to provide feedback on the logical progression of the solution. Taking Fig. \ref{fig:1} as an example, the step-level math correction task finds the error step (Step 4) from the whole solution where the final answer is right.

Despite its potential, step-level automatic math correction faces two main challenges. The first challenge (\textbf{C1}) is that current classification models often focus on superficial patterns between solution steps and final answers, overlooking the underlying reasoning. This limits detailed, accurate feedback. Experiments show that supervised fine-tuning (SFT) LLMs achieve only about 70\% $\mathbf{F_1}$ score in binary classification (Table \ref{table:1}), revealing performance gaps. The second challenge (\textbf{C2}) lies in binary human feedback, which oversimplifies step correctness and ignores partial errors or nuances, as not all steps are entirely right or wrong. A more fine-grained evaluation is necessary to address this complexity.

To address these challenges, we propose \texttt{StepAMC}, a novel RL-LLM approach for step-level automatic math correction. To tackle \textbf{C1}, we use RL to guide the model toward capturing step reasoning rather than relying on shortcuts. A space-constrained policy network further enhances stability and performance. For \textbf{C2}, a fine-grained reward network assigns continuous scores, enabling learning from partial errors and reflecting nuanced human feedback. Experiments show that \texttt{StepAMC} achieves higher accuracy and better aligns with human judgment than existing methods.

The main contributions of this paper are listed as follows.
\begin{itemize}[leftmargin=*, align=left]
    \item We convert step-level automatic math correction for text classification tasks into reinforcement learning settings naturally to boost the reasoning performance of LLMs.
    \item We propose a space-constrained policy network to improve the stability and performance of LLMs and design a fine-grained reward network to give the correctness of each step in a more nuanced manner.
    \item A series of experimental results over two datasets show that our model outperforms the strong baselines, indicating that our \texttt{StepAMC} captures the reasoning behind each step.
\end{itemize}

\section{Related Work}
\subsection{Automatic Homework Correction}
Automatic homework correction aims to check the students' results via artificial intelligence algorithms. 
Several studies are proposed to localize, recognize, and correct multiple-choice problems in tests \cite{supic2014automatic,calado2019application}.
For example, Supic et al. \cite{supic2014automatic} designed a computer vision algorithm to extract handwritten answers from the tests.
Other work has targeted fill-in-the-blank and short-answers questions by detecting and converting the handwritten answers into digital formats \cite{wang2021automatic,li2019bags,burrows2015eras,pulman2005automatic}.
Notably, Tornqvist et al. \cite{tornqvist2023exasag} trained a fine-tuned Transformer-based classifier with explainable outcomes, while Zhang et al. \cite{zhang2022automatic} fine-tuned MathBERT using in-context learning for automatic short answer grading to math questions.

For free-text input, the system analyzes sentence structure and extracts keywords to match the target answer \cite{sukkarieh2003automarking}. A related application is automatic code assessment, where models evaluate code by syntax, performance, and quality \cite{cipriano2022drop,combefis2022automated}. Most relevant is \cite{chaowicharat2023step}, which models solutions as symbolic equation sequences. In contrast, we focus on step-level correction of math solutions expressed in natural language.

\begin{figure}[t!]
    \centering
    \includegraphics[width=0.495\textwidth]{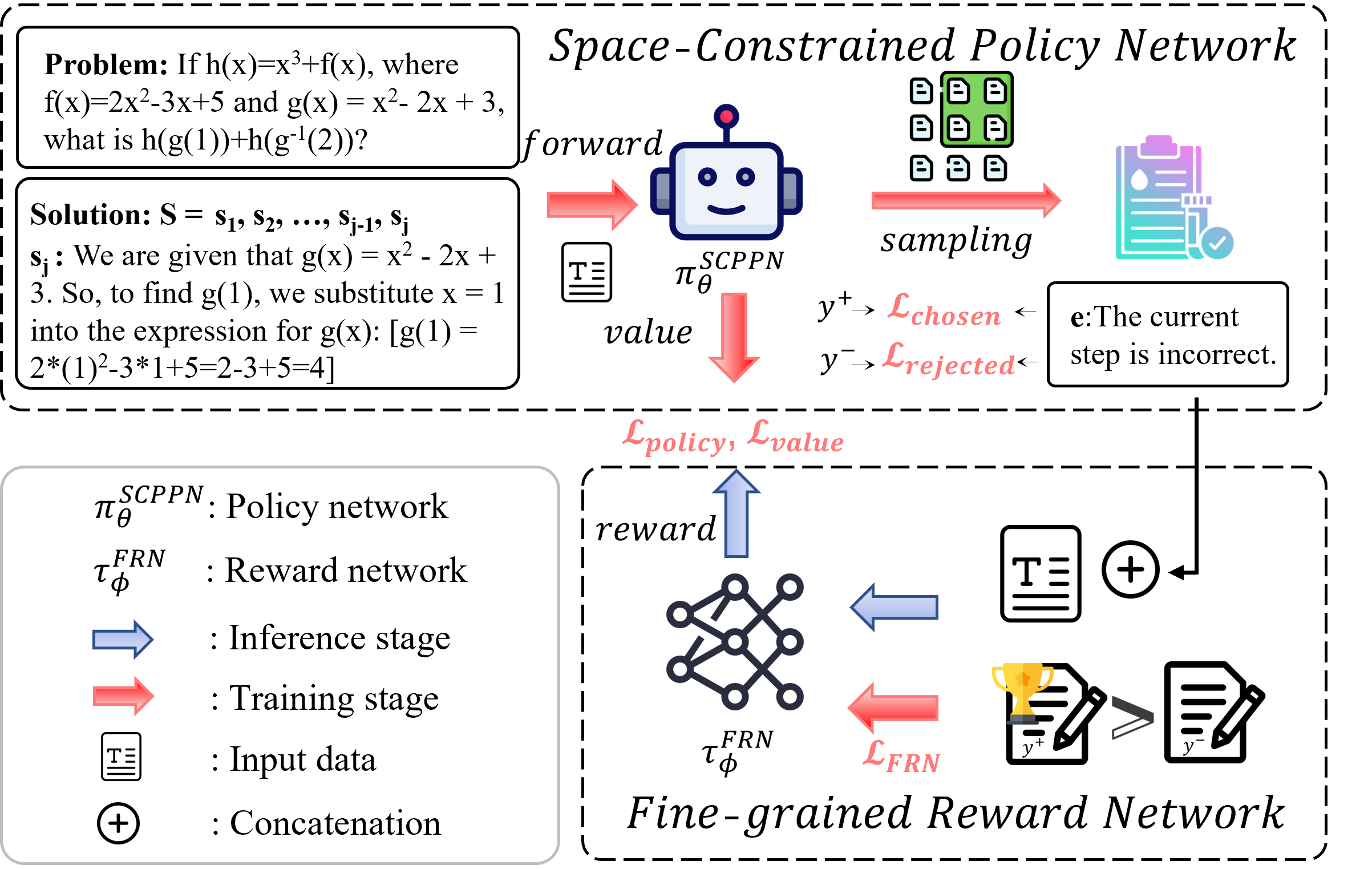}
    \caption{The overall framework of StepAMC.} 
    \label{fig:2} 
\vspace{-4mm}
\end{figure}

\subsection{Artificial Intelligence for Mathematical Reasoning}
Artificial intelligence technologies are widely used for mathematical reasoning~\cite{liu2023mathematical,davies2021advancing,trinh2024solving}. 
Earlier studies mainly phrased the question into symbol representation to infer the final scores \cite{zhang2019gap,huang2017learning}.
Recently, large language models (LLMs) like GPT-4 \cite{openai2023gpt4}, LLaMA \cite{touvron2023llama2} and Qwen \cite{yang2024qwen2} have achieved the state-of-the-art performance in math problem solving, leveraging their strong reasoning capabilities \cite{matzakos2023learning}. 
The most popular algorithm is chain-of-thought (CoT) \cite{wei2022chain}, which solves the math problem step by step. 
Many advanced versions of CoT are proposed, such as tree-of-thought (ToT) \cite{long2023large}, graph-of-thought (GoT) \cite{yao2023beyond}.
Then, tools are used to calculate the results using program and calculators \cite{schick2023toolformer,gou2023tora}.

Ding et al. proposed a Socratic teaching-based LLM for mathematical teaching via conversation \cite{ding2024boosting}.
Furthermore, reinforcement learning strategies are utilized to improve the performance of math reasoning \cite{trung2024reft,kazemnejad24vineppo}.
Unlike these studies that mainly focus on providing the solution of the question, we aim at correcting the steps in solutions given by students.

\section{Our Method}
We propose \texttt{StepAMC}, an RL-based method for step-level automatic math correction, composed of a space-constrained policy network and a fine-grained reward network (Fig. \ref{fig:2}). To address RL instability \cite{henderson2018deep}, we design the space-constrained policy network to improve performance by narrowing the search space. Additionally, we introduce the fine-grained reward network to assess the correctness of each step by converting the binary human feedback into a continual value.

Formally, given a set of problems $D=\{(X_i, Y_i)\}_{i=1}^{|D|}$, where $|D|$ is the number of samples in $D$. For $X=(q, S)$ in each sample of D, $q$ is the question and $S=\{s_1, ..., s_j, ..., s_{|S|}\}$ is the student's solution, where $s_j$ is the $j$-th step of $S$ and $|S|$ is the number of steps. For $Y=\{y_1, ..., y_j, ..., y_{|S|}\}$, $y_j$ is the correctness of step $s_j$. Particularly, $y_j \in \{\text{“correct”, “incorrect”}\}$ depending on the correctness of $s_j$. The goal of this task is to predict $y_j$ based on question $q$ and previous and current steps $\{s_1, ..., s_{j-1}, s_j\}$.  

\subsection{Space-Constrained Policy Network}
We frame step-level automatic math correction as a reinforcement learning problem to understand the reasons behind the solutions.
Reinforcement learning-based optimization in large language models often suffers from instability due to the expansive search space of possible actions. To mitigate this issue, we propose the Space-Constrained Policy Network (SCPN), which reduces the search space by incorporating domain-specific constraints as an auxiliary task. 
Given the context of the question $q$ and previous steps $\{s_1, \ldots, s_{j-1}\}$, the SCPN is trained to predict the correctness of each step $s_j$ by outputting an action $a_j \in \{\text{“correct”}, \text{“incorrect”}\}$. 

Here, we adopt an LLM (e.g., LLaMA, Qwen) as our policy network using LoRA \cite{DBLP:conf/iclr/HuSWALWWC22}. 
Following previous work \cite{trung2024reft}, the response $e$ generated by the policy network can be written by:
\begin{equation}\label{eq:12}
    e = \{a_1, a_2, ... , a_{L-1}, a_L=<eos> \}
\end{equation}
where $L$ is the maximum length. At timestep $t$, the action $a_t$ is sampled from the policy $\pi_{\theta}^\mathrm{SCPN}(\cdot|state_t$) where $a_t$ can be any token in the vocabulary and the state $state_t$ includes all tokens in the input. After each action, the updated state $state_{t+1}$ is is formed by concatenating $state_t$ with $a_t$:
\begin{equation}\label{eq:13}
    state_{t+1} =
    \begin{cases}
        [q,s_{1:j}], & t = 0, \\
        [state_t, a_t], & 1 \leq t \leq L.
    \end{cases}
\end{equation}

\paragraph{Policy Optimization} The SCPN policy optimization leverages the Proximal Policy Optimization (PPO) framework \cite{schulman2017proximal}, a robust reinforcement learning method designed to stabilize training by restricting policy updates. PPO prevents large deviations from the previous policy, ensuring stable and efficient learning. The optimization process incorporates two key components: policy loss and value loss.

The policy loss encourages the policy to favor actions with Generalized Advantage Estimation(GAE) $A_t$ \cite{schulman2015high}, which measures the relative advantage of the action $a_t$:

\begin{equation}\label{eq:3}
     A_t = \delta_t + (\gamma \lambda) \delta_{t+1} + \cdots + (\gamma \lambda)^{L-t} \delta_L,
\end{equation}
where $\delta_t = R(state_t) + \gamma V(state_{t+1}) - V(state_t)$ is the temporal difference, $R(state_t)$ refers to the reward of state $state_t$ which will be discussed in detail in next section, $\gamma$ is the discount factor, and $\lambda$ is a hyperparameter controlling the bias-variance trade-off in advantage estimation. The value function $V(state_{t})$ is implemented as a learnable value head projection in the policy network, which computes the expected cumulative return based on the final hidden state of the input $state_t$. Formally, this can be expressed as:
\begin{equation}
    V(state_t) = \text{ValueHead}(\mathbf{h}_t),
\end{equation}
where $\text{ValueHead}(\cdot)$ is a projection layer, and $\mathbf{h}_t$ is the last hidden state of $state_t$ obtained from the policy network.

To avoid unstable updates, the probability ratio is defined as: 
\begin{equation}\label{eq:4}
    d_t= \frac{\pi^\mathrm{SCPN}_\theta(a_t | state_t)}{\pi^\mathrm{SCPN}_{\theta_{\text{old}}}(a_t | state_t)},
\end{equation}
where $\pi^\mathrm{SCPN}_{\theta_{\text{old}}}$ represents the policy network from the previous iteration, which constrains the update magnitude and stabilizes training. The clipped policy loss is then formulated as:
\begin{equation}\label{eq:5}
    \mathcal{L}_\text{policy} = \mathbb{E}_{e \sim \pi^\mathrm{SCPN}_{\theta_{\text{old}}}} \left[ \min \left( d_t A_t, \text{clip}(d_t, 1 - \epsilon, 1 + \epsilon) A_t \right) \right],
\end{equation}
where $\epsilon$ is a hyperparameter defining the clipping range.

To improve the accuracy of the value prediction, the value loss minimizes the discrepancy between expected return $V(state_t)$ and actual return $R_t$, defined as:
\begin{equation}
\label{eq:7}
\begin{aligned}
&\quad \quad \quad \mathcal{L}_\text{value} = \frac{1}{2} \mathbb{E}_{e \sim \pi^\mathrm{SCPN}_{\theta_{\text{old}}}} [ 
    \max ( \|V(state_t) - R_t\|^2, \\
        &\|\text{clip}(V(state_t), V_{\text{old}}(state_t) - \epsilon, V_{\text{old}}(state_t) + \epsilon) - R_t\|^2 
    ) 
].
\end{aligned}
\end{equation}
where $V_{\text{old}}(state_t)$ represents the value function predicted by the old policy network $\pi^\mathrm{SCPN}_{\theta_{\text{old}}}$ and $R_t = \sum_{t'=t}^L \gamma^{t'-t} R(state_t)$.

The primary policy optimization loss combines the policy loss and the value loss into a single objective function: 
\begin{equation}\label{eq:11} 
    \mathcal{L}_\text{primary} = \mathcal{L}_\text{policy} + c_v \cdot \mathcal{L}_\text{value}, 
\end{equation}
where $c_v$ is a hyperparameter controlling the contribution of the value loss. This formulation balances the policy's optimization of immediate actions with the long-term accuracy of state value predictions.

\paragraph{Policy Constraint}
To further stabilize training and improve the efficiency of policy optimization, we introduce a constraint loss that explicitly regularizes the policy network’s ability to distinguish between correct and incorrect steps. This constraint loss serves as a complementary supervision signal, encouraging the model to prioritize plausible corrections while reducing the likelihood of exploring incorrect actions.

The constraint loss utilizes positive ($y^+$) and negative ($y^-$) labels to guide the policy network in predicting appropriate responses based on the initial state $state_0$. The logits predicted by the policy network are used to compute the probabilities $\pi^\mathrm{SCPN}_\theta(a_t | q; e_{<t})$, which are then optimized using cross-entropy losses. Formally, the constraint loss is defined as:

\begin{equation}\label{eq:chosen_loss}
\mathcal{L}_\text{chosen} = - \mathbb{E}_{e \sim \pi^\mathrm{SCPN}_{\theta_{\text{old}}}}  \sum_{t=1}^L y^+_t\log \pi^\mathrm{SCPN}_\theta(a_t | q; e_{<t}) ,
\end{equation}
\begin{equation}\label{eq:rejected_loss}
\mathcal{L}_\text{rejected} = - \mathbb{E}_{e \sim \pi^\mathrm{SCPN}_{\theta_{\text{old}}}}  \sum_{t=1}^L y^-_t\log \pi^\mathrm{SCPN}_\theta(a_t | q; e_{<t}) ,
\end{equation}
\begin{equation}\label{eq:10}
    \mathcal{L}_\text{const} = \mathcal{L}_\text{chosen} - \mathcal{L}_\text{rejected}.
\end{equation}

The total loss combines the primary and constraint loss: 
\begin{equation}\label{eq:14}
    \mathcal{L}_\text{total} = \alpha \cdot \mathcal{L}_\text{primary} + (1 - \alpha) \cdot \mathcal{L}_\text{const},
\end{equation}
where $\alpha \in [0, 1]$ is a learnable balanced factor.

This integrated objective ensures that the SCPN balances short-term corrections with long-term policy improvements, yielding a robust and effective training process.

\subsection{Fine-grained Reward Network}
In the context of step-level automatic math correction, a critical challenge is to provide nuanced feedback for each step of a student’s solution. Traditional reward signals, often derived from binary human feedback (e.g., correct or incorrect), fail to capture the nuanced differences between partially correct and entirely incorrect steps. To address this, we propose a Fine-grained Reward Network (FRN) that transforms binary feedback into continuous reward values, providing a more detailed evaluation of each step's correctness. Here, we also employ an LLM as our reward network using LoRA.

Given a sample $(X, Y)=((q, \{s_j\}_{j=1}^{|S|}),\{y_j\}_{j=1}^{|S|})$, we construct label pairs $(y_j^+, y_j^-)$, where $y_j^+$ is the original correctness label $y_j$ for step $s_j$, and $y_j^-$ is the inverted label of $y_j^+$. This approach enables the FRN to learn from both positive and negative responses, enhancing the ability to differentiate between correct and incorrect steps. Specifically, the FRN evaluates the correctness of a step by considering both the original input (e.g., the problem statement, prior steps, and the current step) and the generated response by the policy network, formulated as follows:
\begin{equation}\label{eq:1}
    R(state_t) = \tau_{\phi}^\mathrm{FRN}([q;s_{1:j};e_{<t}])
\end{equation}
\noindent where $R(state_t)$ denotes the reward of $state_t$, and $\tau_{\phi}^{FRN}(\cdot)$ is the FRN parameterized by $\phi$.

In the actual training process, we append the fields indicating the positive ($y_j^+$) and negative ($y_j^-$) labels to the input and feed them into the FRN to obtain the reward values $r_j^+=R([q,s_{1:j};y_j^+])$ and $r_j^-=R([q,s_{1:j};y_j^-])$ for the positive and negative sample pairs. We expect the FRN to assign higher reward values to positive samples, which requires satisfying the condition $r_j^+ > r_j^-$ via a pairwise loss.
\begin{equation}\label{eq:2}
    \mathcal{L}_\text{FRN} = - \frac{1}{|D|}\sum_{(X,Y)\in D}\sum_{j=1}^{|S|}\left[ \log \sigma(r_j^+) -  \log \sigma(r_j^-)
    \right]
\end{equation}
\noindent where $\sigma(\cdot)$ is the sigmoid function.

\section{Experiments}
\subsection{Experiments Setups}
\paragraph{Datasets} Due to the lack of real-world datasets containing students' step-by-step solutions, we use two generated datasets, PRM800K \cite{lightman2023let} and Math-Step-DPO-10K \cite{lai2024step}, originally designed to improve the ability of models to solve math problems with step-level annotations, as substitutes.

\textbf{PRM-42K.} We build PRM-42K based on PRM800K, which generates step-by-step solutions via the MATH dataset \cite{hendrycksmath2021} using GPT-4, and then annotates these solutions with human feedback, containing approximately 800,000 step-level annotations categorized into three classes: \{positive, neutral, negative\}. Since our task focuses on assessing the correctness of students' solutions, we treat neutral labels as positive.

We filter out problems labeled as \text{“bad\_problem”} or \text{“give\_up”}, split the solutions into individual steps, shuffle them, and perform random sampling while ensuring that the ratio of positive and negative samples is roughly balanced. 

\textbf{MSD-22K.} We construct MSD-22K by transforming Math-Step-DPO-10K, originally a preference dataset, where each problem is paired with a set of previous solution steps and two candidate next steps: one correct and one incorrect. To adapt this dataset to our task, we transform it into a step-level classification format by combining the problem and previous steps with each of the candidate's next steps. 

Both datasets are divided into training, validation, and test sets in an 8:1:1 ratio (see Table \ref{table:3}). 

\paragraph{Metriccs} To comprehensively evaluate the performance of the proposed model, we employ several metrics: $\mathbf{F_1}$, $\mathbf{Acc}$, $\mathbf{Acc_{pos}}$, and $\mathbf{Acc_{neg}}$, to assess overall classification accuracy and the model's behavior across different label categories.
Specifically, the primary metrics include $\mathbf{F_1}$ and $\mathbf{Acc}$. 
$\mathbf{F_1}$ balances precision and recall, while $\mathbf{Acc}$ represents the proportion of correctly classified samples across all categories. 
Additionally, we introduce $\mathbf{Acc_{pos}}$ and $\mathbf{Acc_{neg}}$ to assess potential bias towards specific classes. For example, $\mathbf{Acc_{pos}}$ is calculated as the ratio of correctly classified positive samples to the total number of positive samples.
\begin{table}[t!]
    \caption{Statistics of PRM-42K and MSD-22K.}
    \label{table:3}
    \centering
    \newcolumntype{"}{@{\hskip\tabcolsep\vrule width 1.2pt\hskip\tabcolsep}}
    \begin{tabular}{p{1.6cm}<{\centering}"
        p{1cm}<{\centering} p{1cm}<{\centering} p{1cm}<{\centering} p{1cm}<{\centering}}
        \toprule[1.2pt] 
        \textbf{Dataset} & \textbf{Train} & \textbf{Val} & \textbf{Test} & \textbf{Total} \\
        \midrule[0.8pt] 
        PRM-42K & 33,248 & 4,354 & 4,156 & 41,758 \\
        MSD-22K & 17,272 & 2,159 & 2,159 & 21,590 \\
        \bottomrule[1.2pt]
    \end{tabular}
\vspace{-4mm}
\end{table}
\paragraph{Baselines} For a more comprehensive assessment, we incorporate two categories of methods as baseline approaches, 
including prompt-based models and fine-tuned models.
For prompt-based models, we select existing LLMs that are good at reasoning, including Qwen-Math-Plus \cite{yang2024qwen2}, DeepSeek-V2.5\footnote{https://huggingface.co/deepseek-ai/DeepSeek-V2.5}, GLM-4-Plus\footnote{https://bigmodel.cn/dev/howuse/glm-4}, GPT-3.5\footnote{https://chatgpt.com}, GPT-4 \cite{openai2023gpt4} and Claude-3.5-Sonnet\footnote{https://www.anthropic.com/news/claude-3-5-sonnet}. 
For fine-tuned models, 1) BERT \cite{devlin2019bert} and RoBERTa \cite{liu2019roberta}, two typical baselines for text classification; 2) SFT, which fine-tunes open-source LLMs (e.g., Qwen-2.5-Instruct); 3) DPO \cite{rafailov2023direct} and PPO \cite{schulman2017proximal}, both of which train using reinforcement learning. 
Note that SFT, DPO, PPO and \texttt{StepAMC} are based on Qwen-2.5-Instruct in Table \ref{table:1}.



\paragraph{Implementation Details} Our training process utilizes eight A800-80GB GPUs and is conducted using the LLaMA-Factory framework\footnote{https://github.com/hiyouga/LLaMA-Factory/tree/main}. For BERT and RoBERTa, we adopt a learning rate of 2e-5, while for SFT, the learning rate is set to 1e-4. In RL-related methods, we initialize training with SFT as a warm-up phase, followed by reinforcement learning with a reduced learning rate of 5e-6. For \texttt{StepAMC}, the hyperparameters are configured as follows: $\gamma = 1$, $\lambda = 0.95$, $\epsilon=0.2$, and $c_v = 0.1$. Across all methods, we consistently train for 3 epochs and report the best-obtained results.

\subsection{Main Results}
\begin{table*}[t]
        \caption{Main results. Imp. means the improvements compared with the strong baseline DPO. The best and suboptimal results are marked in \textbf{bold} and \underline{underlined}.}
        \label{table:1}
	\centering
        \newcolumntype{"}{@{\hskip\tabcolsep\vrule width 1.2pt\hskip\tabcolsep}}
		\begin{tabular}{>{}p{2.6cm}"
        p{1.2cm}<{\centering} p{1.2cm}<{\centering} p{1.2cm}<{\centering} p{1.2cm}<{\centering}"
        p{1.2cm}<{\centering} p{1.2cm}<{\centering} p{1.2cm}<{\centering} p{1.2cm}<{\centering}}
			\toprule[1.2pt] & \multicolumn{4}{c}{\textbf{PRM-42K}} & \multicolumn{4}{c}{\textbf{MSD-22K}}   \\
			\midrule[0.8pt]{\enspace \textbf{Method}}
                & $\mathbf{F_1}$ & $\mathbf{Acc}$ & $\mathbf{Acc_{pos}}$ & $\mathbf{Acc_{neg}}$
                & $\mathbf{F_1}$ & $\mathbf{Acc}$ & $\mathbf{Acc_{pos}}$ & $\mathbf{Acc_{neg}}$ \\
			\midrule[0.8pt] 
			{\enspace \textbf{Qwen-Math-Plus}} 
                & 73.60 & 66.24 & 93.98 & 38.42
                & 71.59 & 66.74 & 83.81 & 49.67 \\
                {\enspace \textbf{DeepSeek-V2.5}}
                & 73.52 & 70.46 & 82.29 & 58.42
                & 73.17 & 67.26 & \textbf{89.30} & 45.21 \\
                {\enspace \textbf{GLM-4-Plus}}
                & 71.17 & 60.77 & \textbf{96.68} & 24.76
                & 67.43 & 68.37 & 65.49 & 71.26 \\
                {\enspace \textbf{GPT-3.5}}
                & 61.53 & 55.56 & 71.08 & 40.04
                & 61.34 & 53.64 & 73.45 & 33.77 \\
		      {\enspace \textbf{GPT-4}}
                & 76.14 & 72.95 & 86.23 & 59.62
                & 76.82 & 75.27 & 81.87 & 68.65 \\
			{\enspace \textbf{Claude-3.5-Sonnet}}
                & 64.75 & 68.07 & \underline{94.03} & 42.11
                & 76.02 & 73.07 & 85.29 & 60.93 \\
                \midrule[0.8pt]
                {\enspace \textbf{$\textrm{BERT}$}}
                & 48.91 & 64.28 & 34.56 & \textbf{93.38}
                & 65.29 & 53.40 & 87.51 & 19.20 \\
                {\enspace
                \textbf{$\textrm{RoBERTa}$}}
                & 69.88  & 63.14  & 85.51  & 40.76 
                & 54.93 & 52.20 & 58.19 & 46.20 \\
                {\enspace \textbf{SFT}}
                & 72.85 & 71.82 & 75.60 & 68.05
                & 72.97 & 67.16 & \underline{88.53} & 45.73 \\
                {\enspace \textbf{PPO}}
                & 73.84 & 74.98  & 70.64 & 79.31
                & 71.89 & 69.43 & 78.08 & 60.76 \\
                {\enspace \textbf{DPO}}
                & \underline{79.28} & \underline{79.43} & 78.73 & 80.13
                & \underline{79.51} & \underline{78.51} & 83.26 & \underline{73.75} \\
                {\enspace \textbf{\texttt{StepAMC} (Ours)}} 
                & \textbf{81.69} & \textbf{81.81} & 81.18 & \underline{82.44}  
                & \textbf{83.09}  & \textbf{82.68} & 85.01 & \textbf{80.33} \\
                {\enspace \scriptsize \textcolor{blue}{Imp.}} 
                & {\scriptsize (\textcolor{blue}{+2.41})} 
                & {\scriptsize (\textcolor{blue}{+2.38})} 
                & {\scriptsize (\textcolor{blue}{+2.45})} 
                & {\scriptsize (\textcolor{blue}{+2.31})}
                & {\scriptsize (\textcolor{blue}{+3.58})} 
                & {\scriptsize (\textcolor{blue}{+4.17})} 
                & {\scriptsize (\textcolor{blue}{+1.75})} 
                & {\scriptsize (\textcolor{blue}{+6.58})} \\
                \midrule[0.8pt]
                {\enspace \quad \textbf{- w/o SCPN}}
                & 76.73 & 77.07 & 75.60 & 78.54
                & 77.21 & 75.31 & 81.96 & 69.57 \\
                {\enspace \quad \textbf{- w/o FRN}}
                & 72.82 & 76.30 & 63.49 & 89.11
                & 68.55 & 71.56 & 61.89 & 81.26 \\
		\bottomrule[1.2pt]
	\end{tabular}
\vspace{-2mm}
\end{table*}
To investigate the effectiveness of \texttt{StepAMC}, we compare it with 11 typical baselines (see Table \ref{table:1}). 
From the results, we obtain the following observations.
\textbf{First}, \texttt{StepAMC} achieves superior performance in both $\mathbf{F_1}$ and $\mathbf{Acc}$ compared to prompt-based and fine-tuned models. A common issue with many baselines is their tendency to predict all samples as either positive or negative, leading to imbalanced $\mathbf{Acc_{pos}}$ and $\mathbf{Acc_{neg}}$. For example, prompt-based methods like GLM-4-Plus achieve high $\mathbf{Acc_{pos}}$ but fail on $\mathbf{Acc_{neg}}$ on PRM-42K, while fine-tuned baselines like BERT show the opposite trend. These results highlight that many models lack the ability to balance predictions across different classes.
\textbf{Second}, fine-tuned baselines like SFT show limited gains, even on LLMs, indicating that simple fine-tuning fails to capture task-specific nuances. In contrast, RL-related methods such as DPO and \texttt{StepAMC} achieve notable performance improvements.
\textbf{Third}, \texttt{StepAMC} consistently outperforms the strong baseline DPO, delivering higher $\mathbf{F_1}$ and $\mathbf{Acc}$ across datasets. However, PPO exhibits poor stability, leading to subpar results and even underperforming SFT on MSD-22K. By reducing search space and providing fine-grained rewards, \texttt{StepAMC} overcomes these challenges, achieving balanced and robust performance.

\subsection{Ablation Studies}
\paragraph{Influence of Main Components}
To evaluate the effectiveness of the main components in \texttt{StepAMC}, we conduct ablation studies (see Table \ref{table:1}). Particularly, we remove the space-constrained policy network (- w/o SCPN) and fine-grained reward network (- w/o FRN). For - w/o SCPN, a general policy network is used, while - w/o FRN replaces the fine-grained reward with binary feedback from humans.

The experimental results demonstrate the great advantages of space-constrained policy network and fine-grained reward network in \texttt{StepAMC}. Removing the SCPN leads to a noticeable drop in both $\mathbf{F_1}$ and $\mathbf{Acc}$, indicating that SCPN plays a crucial role in enhancing stability and guiding the model to comprehend step-level reasoning better. Similarly, replacing the FRN with binary feedback significantly results in a substantial performance decline, particularly in handling nuanced human feedback, as shown by the sharply decreased and highly imbalanced $\mathbf{Acc_{neg}}$ and $\mathbf{Acc_{pos}}$. These findings emphasize the importance of SCPN and FRN in achieving a robust and balanced step-level correction, further validating their contributions to the overall effectiveness of \texttt{StepAMC}.

\paragraph{Influence of Foundation Models}
\begin{table}[t!]
    \caption{Influence of foundation models over PRM-42K.}
    \label{table:2}
    \centering
    \newcolumntype{"}{@{\hskip\tabcolsep\vrule width 1.2pt\hskip\tabcolsep}}
    \begin{tabular}{>{\centering}c"c"
    p{0.8cm}<{\centering} p{0.8cm}<{\centering} p{0.8cm}<{\centering} p{0.8cm}<{\centering}}
        \toprule[1.2pt] 
        \textbf{Backbone} & \textbf{Method} & \textbf{$\mathbf{F_1}$} & \textbf{$\mathbf{Acc}$} & \textbf{$\mathbf{Acc_{pos}}$} & \textbf{$\mathbf{Acc_{neg}}$} \\
        \midrule[0.8pt]
        \multirow{4}{*}{\textbf{Mistral}} 
        & \enspace \textbf{SFT} & 63.94 & 64.82 & 62.73 & 67.28 \\
        & \enspace \textbf{PPO} & 65.24 & 67.20 & 61.55 & 72.86 \\
        & \enspace \textbf{DPO} & 72.21 & 72.33 & 71.90 & 72.76 \\
        & \enspace \textbf{Ours} & 75.72 & 76.18 & 74.30 & 78.06  \\
        \midrule[0.8pt]
        \multirow{4}{*}{\textbf{LLaMA}} 
        & \enspace \textbf{SFT} & 73.50 & 73.27 & 74.16 & 72.38 \\
        & \enspace \textbf{PPO} & 72.71 & 71.99 & 69.35 &  74.64 \\
        & \enspace \textbf{DPO} & 77.15 & 76.78 & 78.39 & 75.17 \\
        & \enspace \textbf{Ours} & 79.91 & 79.64 & 80.99 & 78.30 \\
        \midrule[0.8pt]
        \multirow{4}{*}{\textbf{Qwen}} 
        & \enspace \textbf{SFT} & 72.85 & 71.82 & 75.60 & 68.05 \\
        & \enspace \textbf{PPO} & 73.84 & 74.98  & 70.64 & 79.31 \\
        & \enspace \textbf{DPO} & 79.28 & 79.43 & 78.73 & 80.13 \\
        & \enspace \textbf{Ours} & \textbf{81.69} & \textbf{81.81} & \textbf{81.18} & \textbf{82.44}\\
    \bottomrule[1.2pt]
    \end{tabular}
\vspace{-4mm}
\end{table}

Due to space limitations, we report the results on the PRM-42K dataset here and the conclusions are similar on MSD-22K. As shown in Table~\ref{table:2}, \texttt{StepAMC} consistently outperforms both SFT, PPO and DPO across all foundation models (Mistral-7b-v0.3, Qwen-2.5-7b-Instruct, and LLaMA-3.1-8b-Instruct), achieving higher $\mathbf{F_1}$, $\mathbf{Acc}$, and more balanced class-specific metrics. While the SFT method shows limited ability in fully leveraging the potential of these models, PPO and DPO introduce significant improvements through reinforcement learning. Building on this, \texttt{StepAMC} achieves even greater performance gains, showcasing its adaptability across different backbones. Among the three base models, Qwen-2.5-7b-Instruct yields the best overall results, but \texttt{StepAMC} maintains robust and consistent performance across all foundation models rather than depending on any specific model, underscoring its effectiveness.

\section{Conclusions and Further Work}
In this paper, we proposed \texttt{StepAMC}, a novel reinforcement learning-based large language model for step-level automatic math correction. By introducing a space-constrained policy network and a fine-grained reward network, \texttt{StepAMC} effectively addresses the challenges of capturing the reasoning behind solution steps and incorporating nuanced feedback. Experimental results demonstrate that \texttt{StepAMC} significantly outperforms existing methods, achieving higher accuracy and better alignment with human judgment. Our work highlights the potential of reinforcement learning techniques in enhancing step-level correction and opens new avenues for intelligent educational systems. 
In further work, we would like to explore how to leverage the large-scale math solution dataset to improve the step-level correction.

\section*{Acknowledge}
The authors wish to thank the reviewers for their helpful comments and suggestions.
This research is funded by the National Science and Technology Major Project (No. 2021ZD0114002), the National Nature Science Foundation of China (No. 62477010 and No. 62307028), the Natural Science Foundation of Shanghai (No. 23ZR1441800), Shanghai Science and Technology Innovation Action Plan (No. 24YF2710100 and No. 23YF1426100 ) and Shanghai Special Project to Promote High-quality Industrial Development (No. 2024-GZL-RGZN-02008).

\bibliographystyle{IEEEbib}
\bibliography{main}

\end{document}